%% file: main.tex
\documentclass[10pt,twocolumn,letterpaper]{article}

\input{include/packages}
\input{include/macros}

\usepackage[pagebackref=true,breaklinks=true,letterpaper=true,colorlinks,bookmarks=false]{hyperref}

\iccvfinalcopy % *** Uncomment this line for the final submission

\def\iccvPaperID{7037} % *** Enter the ICCV Paper ID here

\ificcvfinal\fi

\begin{document}

%%%%%%%%% TITLE
\title{
% Grounding by Separation for Detector-Free Weakly Supervised Grounding
Detector-Free Weakly Supervised Grounding by Separation
}

\input{include/authors}

\maketitle
\ificcvfinal\thispagestyle{empty}\fi

%%%%%%%%% ABSTRACT
\begin{abstract}
\input{files/abstract}
\end{abstract}

%%%%%%%%% BODY TEXT

\input{files/intro}
\input{files/related}
\input{files/method}
\input{files/experiments}
\input{files/conclusions}
\clearpage
{\small
\bibliographystyle{include/template/ieee_fullname}
\bibliography{include/gbsbib}
}

\end{document}

%% file: include/packages.tex
\usepackage{include/template/iccv}
\usepackage{times}
\usepackage{epsfig}
\usepackage{graphicx}
\usepackage{amsmath}
\usepackage{amssymb}
\usepackage{booktabs}
\usepackage{multirow}
\usepackage{bbm}
\usepackage[normalem]{ulem}

%% file: include/macros.tex
\def\ours{{GbS}}
\def\oursspace{{GbS }}
\def\oursfull{Grounding by Separation }

\def\ourstask{DF-WSG}
\def\ourstaskspace{DF-WSG }
\def\ourstaskfull{Detector-Free WSG }
\def\ittlossfull{direct image-to-text matching}

\newcommand\secvspace{\vspace{-0.0cm}}
\newcommand\figvspace{\vspace{-0.2cm}}

%% file: include/authors.tex
\author{
    Assaf Arbelle\thanks{Equal contribution}\hspace{-1pt}*$^{1}$,
    Sivan Doveh*$^{1,4}$,
    Amit Alfassy*$^{1,6}$,
    Joseph Shtok$^{1}$,
    Guy Lev$^{1}$,
    Eli Schwartz$^{1,3}$,\\
    Hilde Kuehne$^{1}$,
    Hila Barak Levi$^{4}$,
    Prasanna Sattigeri$^{1}$,
    Rameswar Panda$^{1,2}$,
    Chun-Fu Chen$^{1,2}$,\\
    Alex Bronstein$^{6}$,
    Kate Saenko$^{2,5}$,
    Shimon Ullman$^{4}$,
    Raja Giryes$^{3}$,
    Rogerio Feris$^{1,2}$,
    Leonid Karlinsky$^{1}$\\
    {\tt\small IBM Research$^{1}$, MIT-IBM Watson AI Lab$^{2}$, Tel-Aviv University$^{3}$,} \\
    {\tt\small  Weizmann Institute of Science$^{4}$, Boston University$^{5}$, Technion$^{6}$}
}

%% file: files/abstract.tex
Nowadays, there is an abundance of data involving images and surrounding free-form text weakly corresponding to those images. Weakly Supervised phrase-Grounding (WSG) deals with the task of using this data to learn to localize (or to ground) arbitrary text phrases in images without any additional annotations. However, most recent SotA methods for WSG assume an existence of a pre-trained object detector, relying  on it to produce the ROIs for localization.
In this work, we focus on the task of \ourstaskfull{} (\ourstask{}) to solve WSG without relying on a pre-trained detector. We directly learn everything from the images and associated free-form text pairs, thus potentially gaining advantage on the categories unsupported by the detector.
The key idea behind our proposed \oursfull{} (\ours) method is synthesizing `text to image-regions' associations by random alpha-blending of arbitrary image pairs and using the corresponding texts of the pair as conditions to recover the alpha map from the blended image via a segmentation network. At test time, this allows using the query phrase as a condition for a non-blended query image, thus interpreting the test image as a composition of a region corresponding to the phrase and the complement region. Using this approach we demonstrate a significant accuracy improvement, of up to $8.5\%$ over previous \ourstask{} SotA, for a range of benchmarks including Flickr30K, Visual Genome, and ReferIt, as well as a significant complementary improvement (above $7\%$) over the detector-based approaches for WSG.

%% file: files/intro.tex
\begin{figure}
\begin{center}
% \fbox{\rule{0pt}{2in}\rule{.9\linewidth}{0pt}}
% \hspace{-0.8cm}
% \includegraphics[width=1\columnwidth]{figures/Figure1.pdf}
\includegraphics[width=\columnwidth]{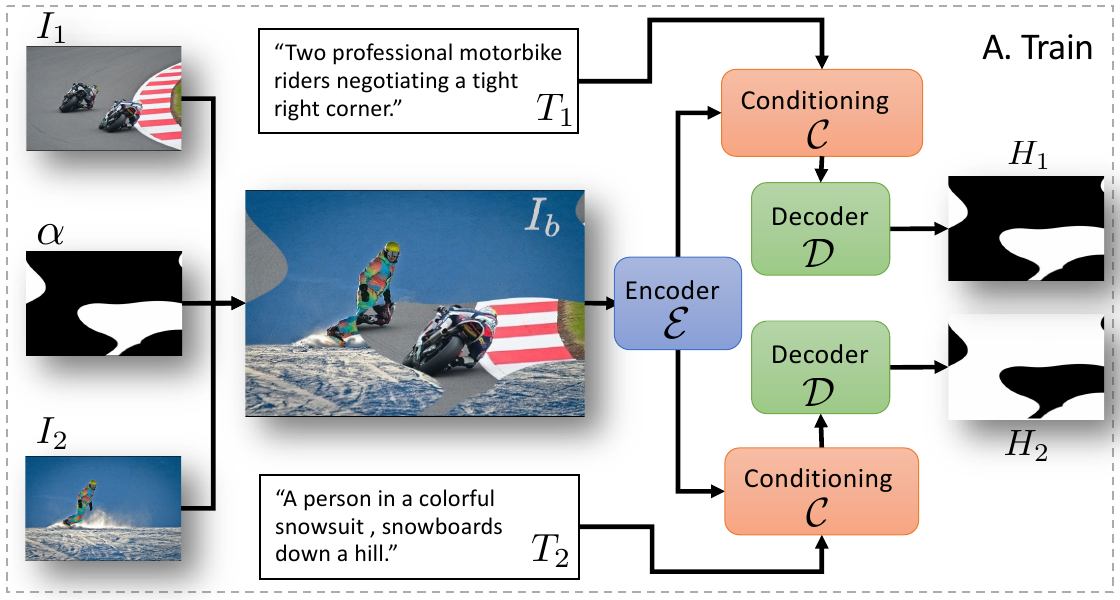}\\
\includegraphics[width=\columnwidth]{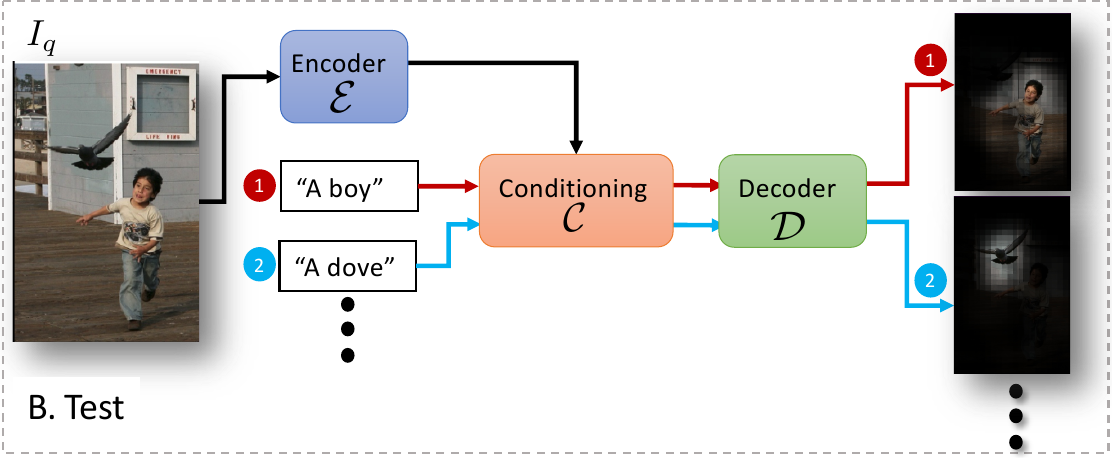}
\end{center}
   \caption{Illustration of our compositional approach. (a) The model is trained to decompose random alpha-blendings of pairs of images conditioned on their associated texts; (b) At test time, the model interprets any image as a composition of two image regions, related and unrelated to the conditioning query phrase, thus grounding the phrase to the image pixels. }
\figvspace
\label{fig:approach_intro}
\end{figure}
\secvspace
\section{Introduction}\label{sec:intro}
\secvspace
As multi-modal text + images data sources become abundant, so grows the importance of natural free-form text supervision \cite{radford2021learning} over the more traditional image labels or image bounding boxes annotation methods. Such multi-modal data (i.e. image-text pairs) can be almost effortlessly and autonomously collected from web pages and documents with illustrations, user captioned personal photos, transcribed videos, and many more. However, such form of automatic supervision poses significant challenges for learning. First, it is noisy in a sense that some of the text words are not relevant to the image; second, it is not well localized in a sense that it is unknown which parts of the image correspond to which parts of the text. In contrast, in traditional annotation the training signal is highly localized: isolated and cropped object images are commonly used in classification, and bounding boxes or polygons around the objects are provided to train detection and/or segmentation models. However, these annotations are commonly manual and are costly to collect.

The above discussion highlights the importance of weakly (and autonomously) supervised multi-modal (images + text) learning in general, and Weakly Supervised Grounding (WSG) in particular. 
In WSG, the model is expected to learn to localize (highlight) image regions corresponding to text phrases. In a sense, WSG is a detection task where the traditional `noun object labels' are replaced by an unbounded set of things describable using natural language. Moreover, the WSG model is expected to learn from image + free-form corresponding text (e.g. caption) pairs without any annotations for correspondence of text words or phrases to image regions.

While earlier WSG methods \cite{akbari2019multi, javed2018learning, xiao2017weakly, zhang2018top} were 'detector-free', all the more recent state-of-the-art (SotA) methods rely on the existence of pre-trained object detectors being the source of the localization RoIs for grounding \cite{datta2019align2ground,gupta2020contrastive,wang2019phrase,chen2020counterfactual,chen2020uniter,lu202012}. Although this 'detector-based' setup benefits from higher performance compared to \ourstaskfull (\ourstask) methods, in a sense it shifts away from the true WSG, as the detector is trained using bounding boxes (which are forbidden in WSG). The use of a detector is indeed plausible when the set of objects supported by the detector significantly overlaps the set of objects (nouns or their taxonomy siblings) appearing in the WSG texts. However, if we need to train for WSG in a different domain (e.g. news~\cite{liu2020visualnews} or technical documents) or for a significantly different set of objects, we are likely to be required to collect a large set of bounding boxes to train a new detector. Experimental evidence for this appears, for example, in a recent detector-based WSG work \cite{datta2019align2ground}\footnote{please see the footnote on page 6 in \cite{datta2019align2ground}}, where it was noted that using the $80$-categories COCO-trained detector for the Flickr30K and Visual Genome (VG) WSG benchmarks performs poorly, as opposed to their best WSG result obtained with the VG trained detector that supports many more relevant categories. 

In this work, we propose an approach for WSG that does not rely on pre-trained detectors and thus addresses the \ourstask{} task. Our approach is based on the idea of image and text compositionality. Having an image + corresponding text pair, we can consider the image as a composition of image regions glued together (like puzzle pieces) to form the whole image, each corresponding to a phrase of the text.
While for a given single image + text pair the composition parts are not known (due to the WSG setting), we can easily simulate a more complex composition by comprising it from any two random image + text pairs. To do so, we can blend the images of the two pairs using a random alpha map $\alpha$, thus making the respective texts of the pairs correspond to the known $\alpha$ and $1 - \alpha$ mapped complementary regions of the blended image. In this way, we can create a reliable localized synthetic training signal for the \ourstask{} model that learns to perform text grounding by learning to separate the blended image to its $\alpha$-mapped constituents conditioned on the respective texts (Figure \ref{fig:approach_intro}a). 
%
% In addition to the separation loss, we further propose two important regularization loss terms which help to prevent the model from learning blending artifacts (thus reducing performance when applied to non-blended test images), as well as to prevent the model from making incorrect references.
%
% We further propose two important regularization loss terms that help preventing the model from learning blending artifacts as well as prevent the model from making incorrect references.
%
At test time, we can apply the trained model on a non-blended query image, which when conditioned on the query phrase is expected to decompose the image to constituents related and not related to the conditioning text (Figure \ref{fig:approach_intro}b).
In addition to the separation loss, we further propose two regularization loss terms which are important for improving the model performance on non-blended test images. These losses help to prevent the model from learning blending artifacts, as well as to prevent the model from making incorrect references.

Our \oursfull{} (\ours) approach obtains a significant, up to $8.5\%$, improvement over previous \ourstask{} SotA \cite{akbari2019multi} for a range of phrase grounding benchmarks including Flickr30K, Visual Genome, and ReferIt. 
Moreover, our performance on these benchmarks is not only comparable to the detector-based WSG SotA \cite{datta2019align2ground,gupta2020contrastive,lu202012}, it is also complementary to them, as our approach is `detector-free' and thus may better support classes that are unknown at the time of detector training. As a result, an ensemble of ours and detector-based SotA methods \cite{gupta2020contrastive,lu202012} improves the Flickr30K detector-based WSG result by over $7\%$, underlining the benefits of our proposed \ours{} approach in situations where a detector is available.
% Moreover, surprisingly, not only that our performance on these benchmarks is comparable to the detector-based WSG SotA \cite{datta2019align2ground, gupta2020contrastive}, it is complementary to these methods (due to handling 'stuff' classes or other classes unknown at detector pre-training) in a way that an ensemble of our and detector based SotA \cite{gupta2020contrastive} methods improves the latter Flickr30K WSG result by over $7\%$, further underlining the benefits of our proposed \ours{} approach.

To summarize, our key contributions are as follows: (i) We propose a novel \ours{} approach for training \ourstask{} models (WSG without assuming a pre-trained detector) based on learning to separate randomly blended images conditioned on the corresponding texts at train time, and applying the learned model on single images with arbitrary text phrase conditioning at test time; (ii) we establish a new SotA for the \ourstask{} task improving significantly the previous best result by up to $8.5\%$ over a range of popular phrase-grounding benchmarks: Flickr30K, VG, and ReferIt; (iii) we provide an extensive ablation study and examine the relative contribution of the components of the \ours{} method; (iv) we obtain a new absolute SotA in WSG on Flickr30K via an ensemble of our \ourstaskspace and the best detector based WSG model, significantly improving the previous SotA result by over $7\%$.
% ; (v) we provide further evidence of wide applicability of the proposed approach by demonstrating SotA results for additional modality mixtures, such as images + recorded speech.

%% file: files/related.tex
\secvspace
\section{Related Work}
\secvspace
Joint analysis of natural images and text is a basic component of many downstream tasks, such as image captioning \cite{cornia2020meshed, karpathy2015deep, pan2020x, vinyals2016show, wang2016image, yao2017boosting, you2016image, zhou2020unified}, text based image retrieval\cite{chen2020uniter,lee2018stacked, lu2019vilbert,lu202012,wang2016learning,young2014image}, visual question answering \cite{alberti2019fusion, antol2015vqa, chen2020counterfactual, gokhale2020vqa, goyal2017making,xu2016ask}, text grounding\cite{akbari2019multi,chen2017msrc,chen2017query,datta2019align2ground,gupta2020contrastive,plummer2015flickr30k,wang2019phrase}, and other general purpose multi-modal learning \cite{huang2020pixel, li2020unicoder,li2019visualbert, li2020weakly,  radford2021learning,  su2019vl, sun2019videobert, tan2019lxmert}. Below we review the text grounding and source separation topics, as the most relevant to our work.

\noindent\textbf{Fully supervised text grounding}.
In the fully supervised grounding setting the training annotations include pairs of phrases and their corresponding image bounding box location. As in object detection, methods employing these detailed annotation train a Region Proposal Network (RPN) to produce image ROIs which are candidates for the grounding target. In \cite{plummer2015flickr30k} joint visual-textual representation space is used for matching the ROIs with the query phrase; instead, \cite{Mao_2016_CVPR} generate text captions for representing each ROI; finally \cite{chen2017msrc} and \cite{chen2017query} slightly modify the task and leverage additional 'context' phrases describing parts of image unrelated to the query phrase, using them too for ROI matching.

\noindent\textbf{Detector-based WSG}.
Most recent methods for WSG (requiring only free-form text captions as image level annotations) assume the availability of a pre-trained object detector, which performs the ROI localization. These methods generally aim to create a joint visual-textual representation space, thus transforming the grounding task into a retrieval task: find the ROI whose embedding best matches the query phrase embedding. 
In \cite{datta2019align2ground} cosine similarity between ROI embeddings and image caption embedding is maximized directly;
\cite{gupta2020contrastive} generate negative text samples using linguistic tools and employ them in a contrastive learning objective between ROIs and the (positive) caption;
\cite{wang2019phrase} match the query phrase to ROI labels produced by multiple pre-trained object detectors;
finally, in a growing body of literature \cite{chen2020uniter, huang2020pixel, li2020unicoder, li2019visualbert, li2020weakly, lu2019vilbert, lu202012, su2019vl, sun2019videobert, tan2019lxmert, zhou2020unified} transformers are used for learning task-agnostic visual-textual representation space where grounding is implemented via retrieval of detector generated image ROIs closest to the query phrase.
%
% Datta \etal~\cite{datta2019align2ground} directly aim to maximize the cosine similarity of the ROI embeddings with the caption embeddings, while Gupta \etal~\cite{gupta2020contrastive} introduce a contrastive loss and by generating negative examples and aiming to maximize the mutual information between the positive pairs of ROI-caption embeddings.  
% Wang \etal~\cite{wang2019phrase} propose an un-paired training regime where the objects multiple pre-trained object detectors are used to extract object proposals, and the object label, as predicted by the detector, is compared to the query phrase. 
% Recent transformer based methods aim to create a task-agnostic, joint representation space for images and text \cite{chen2020uniter, huang2020pixel, li2020unicoder, li2019visualbert, li2020weakly, lu2019vilbert, lu202012, su2019vl, sun2019videobert, tan2019lxmert, zhou2020unified}. These methods are evaluated on many multi-modal tasks while the application of these methods to phrase grounding is done using a pre-trained detector.

\noindent\textbf{Detector-Free WSG (\ourstask)}.
As opposed to detector-based WSG methods, \ourstask{} methods perform dense localization for a given query phrase, thus generating attention heatmaps as opposed to ranking ROIs. The ``Pointing Game" accuracy measure \cite{zhang2016top} is commonly used for \ourstask{} evaluation. Lacking any localization information, \ourstask{} methods often define and optimize some auxiliary task on the weakly supervised data. While the auxiliary task is not identical to the grounding objective, optimization of the task leads to the desired phrase grounding results.
In \cite{xiao2017weakly} joint text and image parsing is employed to enforce structural similarities between the attended image regions and the text parse-tree;
\cite{javed2018learning} employ an attention mechanism to find the common image region among subsets of images which share a specific concept (noun) in the caption; 
\cite{zhang2018top} employ an image \& video captioning model salience maps with respect to the query phrase. 
%This method does not directly train the model to perform language grounding but rather evaluate
The current \ourstask{} SotA \cite{akbari2019multi}, maximizes the likelihood of the caption words in a distribution of image features collected at multiple network depths (scales), as well as optimizing the likelihood of image features in a distribution defined by words in the learned (shared) embedding space. 
% Our work is an \ourstask{} methods which presents a novel auxiliary task of image separation. We further show the importance of regularization terms which support the auxiliary task and align the outcome to the desired phrase grounding objective.

\noindent\textbf{Source-separation methods}. Our approach to \ourstask{} can be considered as doing source-separation, as we separate a randomly generated blended image to its original image sources (conditioned on the texts).  Unconditioned source-separation has been explored extensively using the classical vision methods \cite{be2008blind, gai2011blind, hyvarinen2000independent, levin2007user, levin2002learning, pham1997blind}. Recently, approaches using audio-visual cues has been proposed for the separation of speakers \cite{chao2016speaker, ephrat2018looking, lu2019audio, Owens_2018_ECCV}, musical instruments \cite{ gan2020music, gao2019co,lluis2021music, zhao2019sound, zhao2018sound}, and general sounds \cite{gao2018learning, rouditchenko2019self, xu2019recursive}. Additionally, MixUp \cite{zhang2018mixup} proposed random image blending for augmentation, and unconditioned visual source separation has been examined in \cite{gandelsman2019double,jayaram2020source,Ma_2019_ICCV, lee2018generative,   halperin2019neural,Zou_2020_CVPR, zou2021adversarial}. 
To the best of our knowledge, no previous work has employed (text conditioned) source separation as an objective for learning to perform the text grounding task, as well as for text driven attention in general.
% Our work, on the other hand, performs visual source separation conditioned on caption ques during training on a “synthetic image-caption cocktail dataset”. The conditioning operation on train time allows us to implicitly learn (objects, labels) information used to visually grounding language concepts at test time (in complementary to other existing methods).

%% file: files/method.tex
\begin{figure*}[t!]
\begin{center}
\includegraphics[width=0.9\textwidth]{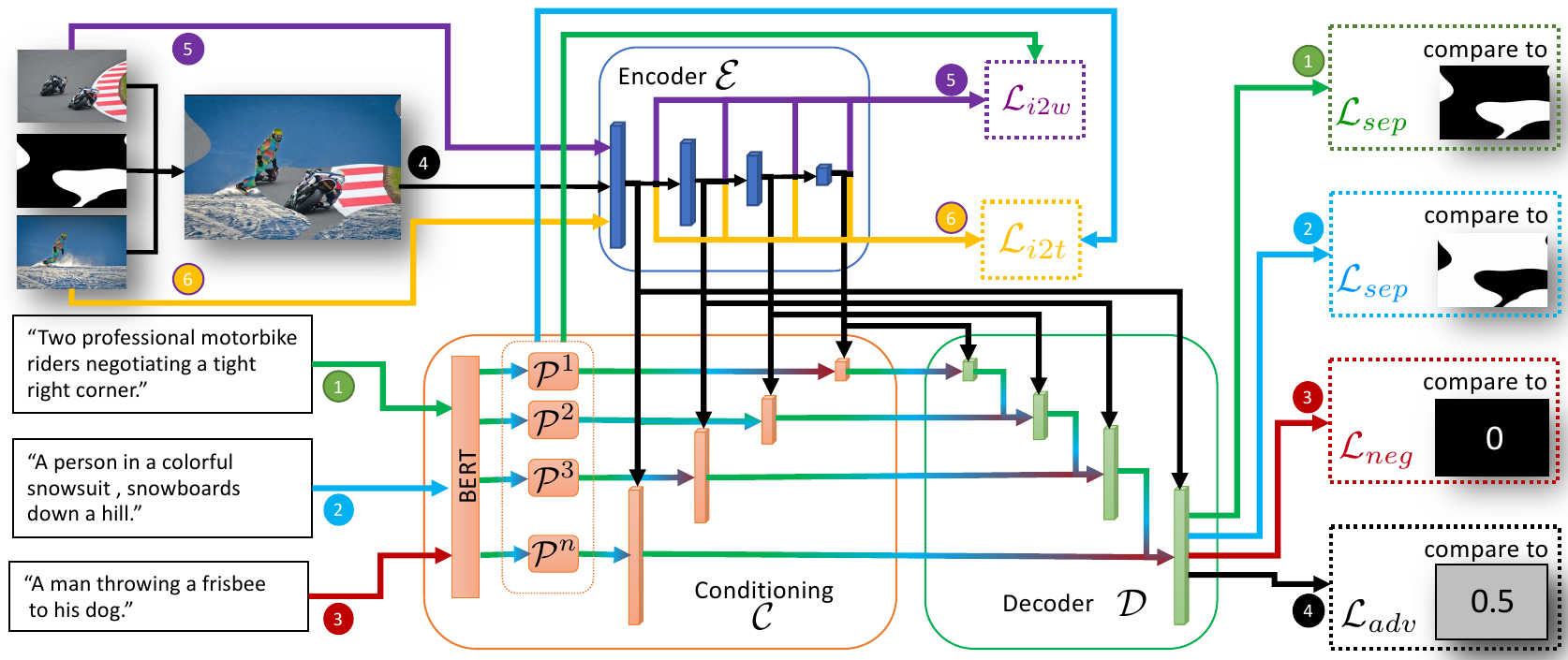} %
%\fbox{\rule{0pt}{2in} \rule{.9\linewidth}{0pt}}
\end{center}
   \caption{Detailed illustration of our model components and flow. Colored and numbered lines represent the flow of different inputs to the respective loss terms. The blended image (black line) flow from $\mathcal{E}$ to $\mathcal{D}$ on the way to the $\mathcal{L}_{adv}$ loss is direct and does not pass through~$\mathcal{C}$.}
\figvspace
\label{fig:model_details}
\end{figure*}
\secvspace
\section{Method}\label{sec:method}
\secvspace
Let $P_1 = (I_1, T_1)$ and $P_2 = (I_2, T_2)$ be two random image ($I_{\times}$) + text ($T_{\times}$) pairs from the \ourstaskspace task training data.
Assume w.l.o.g. that the images are of the same size ($|I_1| = |I_2|$) and let $\alpha$ be a random alpha-map of this size: $|\alpha| = |I_1|$ and $\alpha = \{0 \le \alpha_{i,j} \le 1 | 1 \le i,j \le |I_1|\}$.
Let the blended image $I_b = \alpha \cdot I_1 + (1 - \alpha) \cdot I_2$ be a per-pixel convex combination of $I_1$ and $I_2$, and let $\mathcal{M}(\mathcal{I}, \mathcal{T}) = \mathcal{H}$ be the \oursspace model we would like to train for the \ourstaskspace task, accepting an image $\mathcal{I}$ and text $\mathcal{T}$ as corresponding inputs and returning an output heatmap $\mathcal{H}$. This heatmap $\mathcal{H}$ is predicting the probability of each pixel of the image $\mathcal{I}$ to be related to the text $\mathcal{T}$, in a sense that the pixel belongs to the part of the image described by the text. Our idea is that while the linkage between text parts of $T_1$ and $T_2$ and the corresponding image regions of $I_1$ and $I_2$ is not known (due to the WSG setting), the association between the $T_1$ and $T_2$ components of the concatenated text $T_b = T_1 + T_2$ 
% \hila{confusing - is the text concatenated? the generated synthetic annotations are $(I_b, T_1, \alpha)$ and $(I_b, T_2, 1-\alpha)$} 
and the pixels of the blended image $I_b$ (in a generated `synthetic' pair $\left( I_b, T_b \right)$) is given by construction 
%
% \js{While parts of the input texts $T_1$ and $T_2$ are not localized in their corresponding images $I_1$, and $I_2$, the association of these texts themselves to pixels of the blended image $I_b$ is known by the construction of $I_b$ } 
%
and can be used as a \textit{synthetic} training signal for $\mathcal{M}$. Following this intuition we define our proposed \oursspace main objective (loss):
\begin{equation}
    \mathcal{L}_{sep} = MSE(\mathcal{M}(I_b, T_1), \alpha) + MSE(\mathcal{M}(I_b, T_2), 1 - \alpha)
\end{equation}
where $MSE(x,y) = \frac{1}{|I_b|} \cdot \sum_{i,j}{(x_{i,j} - y_{i,j})^2}$ is the mean-square-error. In this formulation, the model $\mathcal{M}$ is learning to `separate' the blended image $I_b$ conditioned on the text. As mentioned above, any natural image can also be considered as an alpha blending of regions with different semantic meaning (e.g. an overlay of object segments, etc). According to this intuition, our goal is that following training, when provided with a random test image $I_t$ and some corresponding query text $T_q$, computing $\mathcal{M}(I_t, T_q)$ would produce a heatmap $H_q$ such that $I_t$ could be considered as a result of a alpha-blending 
% \js{of two images with $H_q$ as alpha-map: $I_q$, corresponding entirely to the $T_q$ text, and the complement image $\hat{I_q}$.} 
with $H_q$ alpha-map between an image $I_q$ corresponding entirely to $T_q$ and the complement image $\hat{I_q}$ containing everything on $I_t$ that is unrelated to $T_q$:
\begin{equation}
I_t = H_q \cdot I_q + (1 - H_q) \cdot \hat{I_q}
\end{equation}

In the following sections, we provide the architecture specifics of the model we used in our experiments, as well as several additional regularization losses, namely $\mathcal{L}_{adv}$, $\mathcal{L}_{neg}$, and $\mathcal{L}_{i2t}$, which are introduced in sections \ref{sec:adv}, \ref{sec:neg}, and \ref{sec:img_text_align} respectively, and are instrumental to make the proposed construction work well in practice. 
Our overall loss $\mathcal{L}_{\ours}$ is the weighted sum of all of these losses:
\begin{equation}
    \mathcal{L}_{\ours} = 
    \mathcal{L}_{sep} 
    + \gamma_{adv}\cdot\mathcal{L}_{adv}
    + \gamma_{neg}\cdot\mathcal{L}_{neg}
    + \gamma_{i2t}\cdot\mathcal{L}_{i2t}
\end{equation}
An extensive ablation study examining our design choices is provided in section \ref{sec:ablations}. The details of our \oursspace model are illustrated in Figure \ref{fig:model_details}.

\secvspace
\subsection{Model and text conditioning architecture}\label{sec:arch}
\secvspace
Our \oursspace model $\mathcal{M}(\mathcal{I}, \mathcal{T}) = \mathcal{H}$ is comprised of an encoder $\mathcal{E}(\mathcal{I})=E$, a text conditioning module $\mathcal{C}(E, \mathcal{T})=C$, and a decoder $\mathcal{D}(C) = \mathcal{H}$ returning the final output:
\begin{equation}
    \mathcal{H} = \mathcal{M}(\mathcal{I}, \mathcal{T}) = \mathcal{D}(\mathcal{C}(\mathcal{E}(\mathcal{I}), \mathcal{T}))
\end{equation}

\noindent\textbf{The encoder} $\mathcal{E}(\mathcal{I})=E$ is comprised of several (CNN) blocks with (stride $r$) pooling layers in between the blocks. We set $E = [E^1,\ldots,E^n]$ to be a list of tensor outputs of $n$ last blocks ordered in such a way that $E^1$ is the output of the last block (note that $|E^{i+1}| = r \cdot |E^i|$ due to pooling stride). 

\noindent\textbf{The text conditioning module} $\mathcal{C}(E, \mathcal{T})=C$ is comprised of: (i) a text embedding model (e.g. 
% ELMO \cite{peters2018deep} or 
BERT \cite{devlin2018bert}) $\mathcal{N}(\mathcal{T}) = [W_1, ..., W_s]$ returning a list of word embeddings (in the context of the full text $\mathcal{T}$); followed by (ii) a projection module $\mathcal{P}^i(W_j) = W_j^i$, for each encoder $\mathcal{E}$ block $i \in [1,...,n]$, adapting the word embeddings to the space of visual features of $E^i$; followed by (iii) averaging over the words $W^i = \frac{1}{s}\sum_{j}{W_j^i}$ to obtain the full text $\mathcal{T}$ embedding (again per block); and finally followed by (iv) the text attenuation module: 
\begin{equation}
    \mathcal{A}(E^i,W^i) = \exp\left(-\left|\frac{E^i}{||E^i||_2} - \frac{W^i}{||W^i||_2}\right|\right) \cdot E^i = C^i
\label{eq:atten}
\end{equation}
where all operations are element-wise, $W^i$ is broadcasted to all the spatial locations of the tensor $E^i$, and the attenuation enhances locations of $E^i$ that are closer to the projected text embedding for that block. The output of the text conditioning module is hence the per-block list: $C = [C^1,\ldots, C^n]$.

\noindent\textbf{The decoder module}  $\mathcal{D}(C) = \mathcal{H}$ converts the text attenuated image encoding $C$ into the final predicted heatmap $\mathcal{H}$. It is comprised of a series of ResNet \cite{he2016deep} blocks $[D_1,\ldots,D_n]$ such that the first block $D_1$ gets $C_1$ as the input:~$O_1 = D_1(C_1)$, and similarly to U-Net \cite{ronneberger2015unet} each subsequent block receives a combination of the up-scaled previous output and the input:~$O_i=D_i(cat(C_i, U_r(O_{i-1})))$, where $cat$ is channel-wise concatenation and $U_r$ is the spatial up-scaling by the factor of~$r$. We set the number of channels in $O_i$ same as in $C_i$ except for $O_n=\mathcal{H}$ which is the final output of the decoder $\mathcal{D}$ and has a single channel.

\secvspace
\subsection{Unconditioned adversary loss: $\mathcal{L}_{adv}$}\label{sec:adv}
\secvspace
Naturally, images blended with a random alpha-map $\alpha$ differ from natural images and contain blending artifacts that could in turn be leveraged by the model $\mathcal{M}$ in order to produce the source separation. This is an unwanted behaviour that can increase the model's overfit to training data and decrease test performance. To reduce this effect we introduce an adversarial loss reducing the model's use of parameters that build upon these artifacts:
\begin{equation}
    \mathcal{L}_{adv} = MSE(\mathcal{D}(\mathcal{E}(I_b)), 0.5 \cdot \mathbbm{1}_{|\mathcal{H}|})
\end{equation}
where $0.5 \cdot \mathbbm{1}_{|\mathcal{H}|}$ stands for a uniform heatmap with $0.5$ in all pixels indicating maximally uncertain prediction in case no text conditioning was provided. 
% An interesting future research direction, beyond the scope of this work, is also to explore how the proposed adversarial loss $\mathcal{L}_{adv}$ could be used to learn to generate better than random (perhaps even semantic) alpha maps that in turn would make more natural looking blended images and further advance the \ours{} model performance.

\secvspace
\subsection{Negative texts loss: $\mathcal{L}_{neg}$}\label{sec:neg}
\secvspace
We expect the model not only to produce correct $\alpha$ predictions matching the conditioning on the corresponding texts $T_1$ and $T_2$, but also to learn to `reject' conditioning text that do not match. In other words, given a random unrelated text $T_{neg}$ we want the model to produce close to zero prediction on the blended image $I_b$ indicating that no pixel represents this text. 
% \js{So to summarize, with positive text you expect the model to predict 1.0, with no text it should predict 0.5, and with negative text you expect it to predict 0. Maybe add a few lines pointing out this gradation, and how the losses make up a consistent picture of model`s behaviour.}
We therefore define the negative loss to optimize for this requirement:
\begin{equation}
    \mathcal{L}_{neg} = MSE(\mathcal{D}(\mathcal{C}(\mathcal{E}(I_b),T_{neg})),0 \cdot \mathbbm{1_{|\mathcal{H}|}})
\end{equation}

\secvspace
\subsection{Direct image-to-text alignment: $\mathcal{L}_{i2t}$}\label{sec:img_text_align}
\secvspace
Our conditioning module $\mathcal{C}$ is using an attenuation strategy that is based on the similarity between the visual features $\left\{E^i\right\}$ returned by different depth blocks of the encoder $\mathcal{E}$ and the text embedding $\left\{W^i\right\}$ computed as explained in section \ref{sec:arch}. It is therefore likely that a stronger alignment between the $\left\{W^i\right\}$ and the $\left\{E^i\right\}$ would result in a more meaningful attenuation and in turn improved results. In this light, and inspired by ideas from \cite{akbari2019multi} and \cite{radford2021learning}, we added a direct image-to-text alignment loss $\mathcal{L}_{i2t}$ between (non-blended) batch images and their corresponding batch texts. First we compute a similarity between each pair of image \#$m$ and the text corresponding to (another or same) image
% %
% \js{unclear. We are not talking about blended images anymore? these are similarities between texts and images from different text-image pairs in the batch?} 
% %
\#$k$ in the batch:
\begin{equation}
    Z_{k,m} = \max_{i}\left[\cos\left(\sum_{xy}\left[\cos_+\left(W_k^i, E_m^{i,xy}\right) \cdot E_m^{i,xy}\right], W_k^i\right)\right]
\end{equation}
where $\cos$ denotes the cosine similarity and the $\cos_+$ denotes its positive part,
% (abusing notation we allow it between a vector and a tensor meaning it is performed on the corresponding dimension of the tensor and the vector is broadcasted everywhere else), and broadcasting is used in element-wise multiplication. 
and $E_m^{i,xy}$ indicates the feature vector at spatial location $(x,y)$ in the $E_m^i$ tensor. Then we compute the $\mathcal{L}_{i2t}$ loss as:
\begin{align}
    \mathcal{L}_{i2t} = 
    & \sum_{k} \text{CE} \left[ \text{softmax} \left( t_{i2t} \cdot Z_{k,\boldsymbol{\cdot}} \right), k \right] + \\ 
    & \sum_{m} \text{CE} \left[ \text{softmax} \left( t_{i2t} \cdot Z_{\boldsymbol{\cdot},m} \right), m \right]
\end{align}
where $Z_{k,\boldsymbol{\cdot}}$ and $Z_{\boldsymbol{\cdot},m}$ stand for the text \#$k$ row and image \#$m$ column of the matrix $Z$ respectively, $t_{i2t}$ is the softmax temperature, and $\text{CE}$ is the cross-entropy loss with respect to the index of the `correct answer'. The `correct answer' in this case is the respective row or column index itself, similarly to \cite{radford2021learning} we would like the text to best match its corresponding image in the batch - symmetrically when looking at the set of all batch images or all batch texts. Finally, direct matching of the text to the image also produces a heatmap predicting pixel correspondence to the query text. Therefore, in addition to $\mathcal{H}$ returned by the decoder $\mathcal{D}$ (slightly abusing notation, also referred to as $\mathcal{H}_{\ours}$ below), we define an additional output $\mathcal{H}_{i2t}$ from our model, which is the attention map produced by the direct matching:
\begin{equation}
    \mathcal{H}_{i2t}\left(x,y\right) =\max_{i}\left[U_{|E^n|}\left(\cos_+\left(W_k^i, E_m^{i,xy}\right)\right)\right]
\end{equation}
here $U_{|E^n|}$ up-scales to spatial size of $|E^n|$. In our experiments (Section \ref{sec:experiments}) we found that $\sqrt{\mathcal{H}_{\ours} \cdot \mathcal{H}_{i2t}}$, namely the per pixel geometric mean of $\mathcal{H}_{\ours}$ and $\mathcal{H}_{i2t}$, produces the best result and in the following we consider this geometric mean to be the main output of our \oursspace model $\mathcal{M}$.

\input{tables/results_main}
\input{tables/detector_detailed}
%
% \subsection{Text augmentation}\label{sec:text_aug}
% Being trained with free-form text labels, WSG setting naturally suffers from quite sparse sampling of text labels out of the huge variety of synonymous possibilities. Augmentation is the standard way of battling training sample sparsity and we employ it both on image and text sides in \oursspace model training.... \leonid{Guy - please put stuff here if we will have anything helping by submission, otherwise we can add the dropout 0.5 in implementation details and remove this section}

%% file: tables/results_main.tex
\begin{table*}[t]
\begin{center}
\begin{tabular}{ llcccc }
%  \cline{4-6}
\toprule
 \multirow{2}{*}{Method} & \multirow{2}{*}{Backbone} & \multirow{2}{*}{Training} & \multicolumn{3}{c}{ Test Accuracy}\\
%  &&& \multicolumn{3}{c}{ Test Accuracy}\\
%  \hline

  &  &  & VG & Flickr30K & ReferIt\\ 
 \midrule
 Baseline & Random & - & 11.15 & 27.24 & 24.3 \\  
 Baseline & Center & - & 20.55 & 49.20 & 30.30 \\
 \midrule
 TD \cite{zhang2018top} & Inception-2 & VG & 19.31 & 42.40 & 31.97 \\
 SSS \cite{javed2018learning} & VGG & VG & 30.03 & 49.10 & 39.98 \\
 MG \cite{akbari2019multi} & VGG & VG & 48.76 & 60.08 & \textcolor{red}{\textbf{60.01}} \\
 \oursspace  (ours) & VGG & VG & \textcolor{red}{\textbf{53.40}} & \textcolor{red}{\textbf{70.48}} & 59.44\\
 \midrule
 MG \cite{akbari2019multi} & PNASNet & VG & 55.16 & 67.69 & 61.89 \\
%   \hline
 \oursspace (ours) & PNASNet & VG &\textcolor{blue}{\textbf{55.91}}& \textcolor{blue}{\textbf{73.39}} & \textcolor{blue}{\textbf{62.24}}\\
 \midrule
  \midrule
 \multirow{2}{*}{Method} & \multirow{2}{*}{Backbone} & \multirow{2}{*}{Training} & \multicolumn{3}{c}{ Test Accuracy}\\
 &  &  & VG & Flickr30K & ReferIt\\ 
 \midrule
 FCVC \cite{fang2015captions} & VGG & MS-COCO & 14.03 & 29.03 & 33.52 \\
 VGLS \cite{xiao2017weakly} & VGG &  MS-COCO & 24.40 & - & - \\
 MG \cite{akbari2019multi} & VGG &  MS-COCO & 47.94 & 61.66 & 47.52 \\
 \oursspace  (ours) & VGG &  MS-COCO & \textcolor{red}{\textbf{52.00}} & \textcolor{red}{\textbf{72.60}} & \textcolor{red}{\textbf{56.10}} \\
 \midrule
 MG \cite{akbari2019multi} & PNASNet &  MS-COCO & 52.33 & 69.19 & 48.42 \\
%  \hline
 
 \oursspace  (ours) & PNASNet &  MS-COCO & \textcolor{blue}{\textbf{52.70}} & \textcolor{blue}{\textbf{74.50}} & \textcolor{blue}{\textbf{49.26}}\\
 \midrule
 \oursspace  (ours) ensemble & - &  MS-COCO & \textbf{54.55} & \textbf{75.60} & \textbf{58.21}\\
 \bottomrule
\end{tabular}
\end{center}
\caption{Comparison with the state of the art \ourstask{} methods evaluted using the ``pointing game" accuracy on Visual Genome (VG), Flickr30K, and ReferIt. Our \ours{} method outperforms \ourstask{} SotA when using corresponding backbones (VGG or PNASNet) by up to 10.4\%. In \textcolor{red}{\textbf{red}}: best results with VGG; in \textcolor{blue}{\textbf{blue}}: best results with PNASNet; in \textbf{bold black}: result of ensembling our \ours{} models.
}
\figvspace
\label{tab:main_results}
\end{table*}

%% file: tables/detector_detailed.tex
\begin{table*}[h]
\setlength\tabcolsep{5pt}
\begin{center}
\begin{tabular}{ l|| c |c c c c c c c c c }
% \hline
 Method & Overall &People & Animals & Vehicles & Instruments & Bodyparts & Clothing & Scene & Other \\
 \midrule
  Ours \small{(VGG)} & 72.6 & 82.5 & \textcolor{blue}{\textbf{91.5}} & 81.1 & 56.6 & 34.8 & 58.6 & 70.9 & 59.9 \\ 
  Ours \small{(PNASNet)} & 74.5 & 83.6 & 89.3 & 92.1 & \textcolor{blue}{\textbf{83.3}} & \textcolor{blue}{\textbf{53.2}} & 50.1 & 71.3 & 66.7 \\ 
  Align2Ground~\cite{datta2019align2ground}&71.0&-&-&-&-&-&-&-&-\\
  InfoGround (IG)~\cite{gupta2020contrastive} & \textcolor{blue}{\textbf{76.74}} & 83.2 & 89.7 & 87 & 69.7 & 45.1 & \textcolor{blue}{\textbf{74.5}} & 80.6 & 67.3 \\
  12\small{-in-}1~\cite{lu202012} & 76.4 & \textcolor{blue}{\textbf{85.7}} & 82.7 & \textcolor{blue}{\textbf{95.5}} & 77.4 & 33.3 & 54.6 & \textcolor{blue}{\textbf{80.7}} & \textcolor{blue}{\textbf{70.6}} \\
  \midrule
  \small{IG} + 12\small{-in-}1 &  81.1 & 87.1 & 90.4 & 95.5 & 74.2 & 61.5 & 74.0 & 79.9 & 73.5 \\
  Ours\small{(VGG)} + IG &  83.9 & 88.8 & 96.1 & 93.4 &74.4 & \textbf{65.2} & 77.2 & 82.4 & 76.8 \\
  Ours \small{(PNASNet)} + IG &  83.4 & 87.3 & 95.2 & 95.4 & 75.2 & 62.3 & \textbf{78.3} & 81.5 & 77.2 \\
  Ours \small{(VGG)} + 12\small{-in-}1 &  \textbf{85.9} & \textbf{93.4} & \textbf{97.4} &  96.4 & 79.6 & 52.6 & 78.0 & \textbf{83.6} & 78.3 \\
  Ours \small{(PNASNet)} + 12\small{-in-}1 &  84.9 & 93.3 & 96.5 & \textbf{96.8} & \textbf{81.7} & 54.8 & 71.4 & 82.3 & \textbf{78.9} \\
 
\end{tabular}
\end{center}
\caption{Detailed comparison with detector-based WSG methods \cite{datta2019align2ground, gupta2020contrastive} on Flickr30K. The last two lines are an ensemble of our \oursspace models with the SotA InfoGround (IG) method \cite{gupta2020contrastive}. In \textcolor{blue}{\textbf{blue}} - best single model result, in \textbf{bold black} - best overall result. 12\small{-in-}1~\cite{lu202012} pointing accuracy results computed using official code. Align2Ground~\cite{datta2019align2ground} did not provide detailed results and did not release their code.}
\figvspace
\label{tab:detector_detailed_results}
\end{table*}

%% file: files/experiments.tex
\secvspace
\section{Experiments}\label{sec:experiments}
\secvspace
\input{files/datasets}
\input{files/implementation}

\input{files/results}
\input{files/ablations}

%% file: files/datasets.tex
\subsection{Datasets}\label{sec:datasets}
\secvspace
\noindent\textbf{MS-COCO 2014} \cite{lin2014microsoft} consists of $82,783$ training and $40,504$ validation images. Each image is associated
with five captions describing it.

\noindent\textbf{Flickr30k Entities} \cite{plummer2015flickr30k} is based on Flickr30k \cite{young2014image} and contains $224K$ phrases describing localized bounding boxes in $\sim31K$ images each described by 5 captions. 
For evaluation, we use the same 1k images from the test split as in~\cite{akbari2019multi}.

\noindent\textbf{VisualGenome (VG)} \cite{krishna2017visual} has $77,398$ train, $5000$ validation, and $5000$ test images. Each image comes with a set of free-form text annotated bounding boxes.

\noindent\textbf{ReferIt} has 20,000 images and 99,535 segmented image regions from the IAPR TC-12 \cite{grubinger2006iapr} and the SAIAPR-12 datasets \cite{chen2017query} respectively. Images also have an associated description for the entire image, and the image regions were collected in a two-player game \cite{kazemzadeh2014referitgame} with approximately $130K$ isolated entity descriptions. We use the same $9K$ training, 1k validation, and $10K$ test images split as in \cite{akbari2019multi}.

%% file: files/implementation.tex
\secvspace
\subsection{Implementation Details}\label{sec:impl_details}
\secvspace
% Our code (implemented with PyTorch~\cite{NEURIPS2019_9015}) is provided in supplementary and would be released upon acceptance.
All experiments were conducted on 4 Nvidia V100 GPU machine. We used the VGG~\cite{simonyan2014very} backbone from the torchvision~\cite{marcel2010torchvision} library, the PNASNet~\cite{liu2018progressive} from the TIMM library \cite{rw2019timm}, and BERT~\cite{devlin2018bert} from the huggingface-transformers library \cite{wolf-etal-2020-transformers}.
% , and the ELMO from~\cite{Gardner2017AllenNLP}. 
As in~\cite{akbari2019multi}, the VGG and PNASNet are ImageNet pre-trained.
All experiments, unless otherwise noted, use the following configuration (found using MS-COCO validation set):
(i) training batch of size $8$ (pairs of images and text); 
(ii) half of the batch alpha maps are generated using Perlin noise~\cite{perlin1985image} and half using a combination of two random Gaussians (more details in Sec. \ref{sec:abl_alpha});
(iii) the pre-trained BERT model is frozen;
(iv) the projection modules $\mathcal{P}^i$ are a single fully connected layer;
(v) we use $n=2$ layers for the decoder $\mathcal{D}$ (Section~\ref{sec:arch}); 
(vi) the decoder ResNet blocks $D_i$ have $512$ output planes ($1$ for the final output block) and stride $1$;
(vii) the pooling layers stride is $r=2$; 
(viii) the losses weights are: $\gamma_{i2t}=0.1$, $\gamma_{neg}=1$, $\gamma_{adv}=1$; 
(ix) the softmax temperature is $i_{i2t}=10$;
(x) we use the ADAM optimizer~\cite{kingma2014adam} and a linear LR schedule starting from $LR=0.0001$ and dividing it by 10 every $50K$ steps; 
(xi) we use $50\%$ dropout augmentation for the text, and random crop + $512\times512$ resize, color jitter, horizontal flip, and grayscale augmentations for the images. 
% The images were randomly cropped and resized to $512\times512$ pixels.

%% file: files/results.tex
\secvspace
\subsection{Results}\label{sec:results}
\secvspace
We follow the experimental protocol of \cite{akbari2019multi}, using the same data and splits for training, validation and testing. Specifically, in our experiments we evaluate our approach in two training setups: using either MS-COCO train split or the VG train split for training respectively. In both cases, as in~\cite{akbari2019multi}, the resulting models are evaluated on the test splits of Flickr30K, VG, and ReferIt. Same as~\cite{akbari2019multi}, we report the pointing-game accuracy~\cite{zhang2016top} as our performance estimate in all of the experiments. Specifically, for the set of test `image + query phrase' pairs, we report the percent of pairs for which the maximal point of the predicted heatmap for the pair was inside the ground truth annotation bounding box.

The results of our \ours{} approach evaluation and comparison to other \ourstask{} works (not using pre-trained detectors according to the definition of \ourstask{}) are provided in Table \ref{tab:main_results}. As we can see from the table, in both training regimes our \ours{} models significantly outperform the previous best results on all benchmarks using the matching backbones with $4-10.9\%$ absolute improvements for the lighter VGG backbone and $0.4-5.7\%$ absolute improvement for the much heavier PNASNet backbone. More specifically, we observe significant over $5.7\%$ gains on Flickr30K in all training regimes, over $7\%$ gains in ReferIt under MS-COCO training, and over $4\%$ gains on VG in all training regimes using he VGG backbone. 
% Relatively small improvement on VG using PNASNet in both training regimes can likely be attributed to the very large size of the network making it slower to train.
\input{figures/qualitative}

Interestingly, we also found that our proposed \oursspace approach is in fact \textit{complementary} to the detector based WSG methods and can be effectively used to boost their performance. Any detector based method output can be converted to a heatmap by simple assignment of the bounding box scores to pixels of the bounding box (e.g. taking max for overlaps). As we show in Table \ref{tab:detector_detailed_results}, a simple geometric average between the heatmap produced by our model and the heatmap resulting from the best performing detector based methods significantly boosts the pointing game accuracy of the latter indicating our model has learned to produce complementary predictions (e.g. for object categories less supported by the detector, such as some of the instruments and body parts in Table \ref{tab:detector_detailed_results}) boosting the combined performance by $\ge 7\%$ even over the detector-based SotA WSG approaches \cite{datta2019align2ground, gupta2020contrastive,lu202012}. 
Additionally, we include a comparison to an ensemble of the two SotA detector-based methods InfoGround (IG)~\cite{gupta2020contrastive} and 12-in-1~\cite{lu202012} without our \oursspace model ('IG + 12-in-1' line in the table). As can be seen, an ensemble of our \oursspace model with any of these detector-based methods performs significantly better than an ensemble of the detector-based models between themselves (with the gain of $2.8\%$ and $4.8\%$ respectively). This shows that the gains obtained using an ensemble with \oursspace are not simply due to a combination of models, but rather likely stem from the \oursspace model being truly complementary to the detector-based methods.
Some qualitative examples illustrating situations when not relying on the (more constrained) vocabulary of a pre-trained detector helps the grounding task are provided in Figure \ref{fig:qualitative}.

%% file: figures/qualitative.tex
\begin{figure*}[t!]
\begin{center}
\def\figh{0.9in}
\setlength\tabcolsep{0.4pt}
\begin{tabular}{cccccc|c}
\raisebox{2\normalbaselineskip}[0pt][0pt]{\rotatebox{90}{\ours{}}}&
\includegraphics[height=\figh]{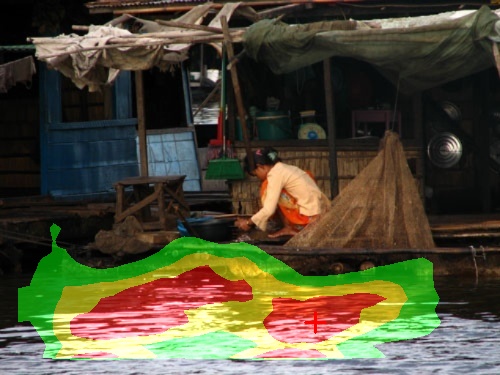}&
\includegraphics[height=\figh]{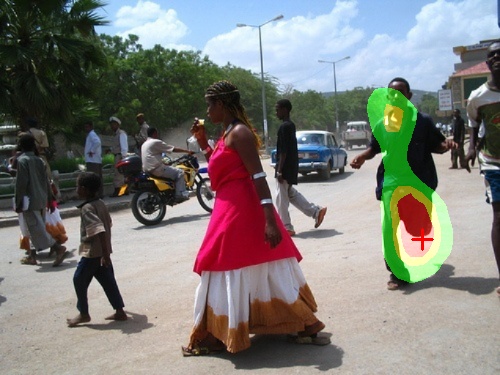}&
\includegraphics[height=\figh]{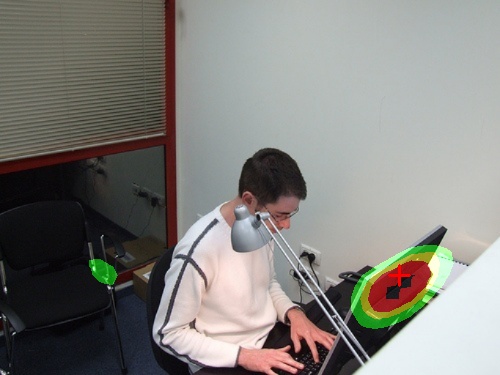}&
\includegraphics[height=\figh]{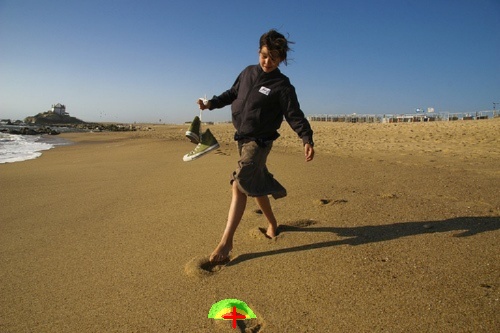}&
\includegraphics[height=\figh]{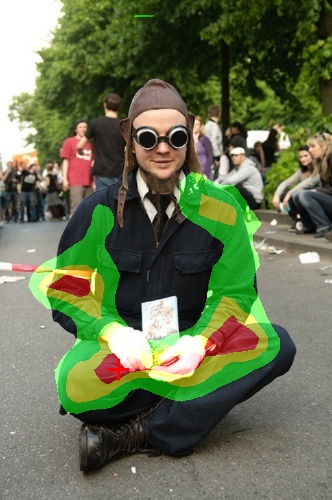}&
\includegraphics[height=\figh]{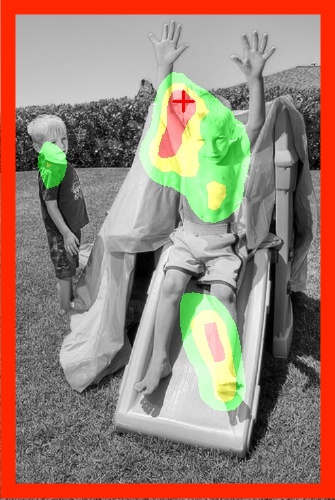}\\
~&\small{``The water"}&\small{``A man running"}&\small{``Desktop"}&\small{``Footprints"}&~~~\small{``His legs"}~~~&~~\small{``Another~kid"}\\
\raisebox{2\normalbaselineskip}[0pt][0pt]{\rotatebox{90}{IG\cite{gupta2020contrastive}}}&
\includegraphics[height=\figh]{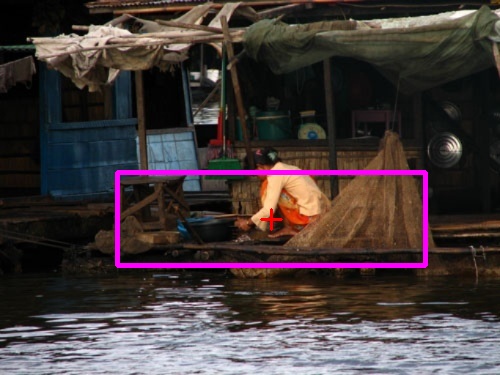}&
\includegraphics[height=\figh]{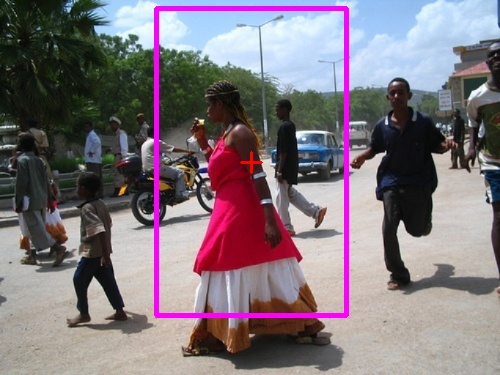}&
\includegraphics[height=\figh]{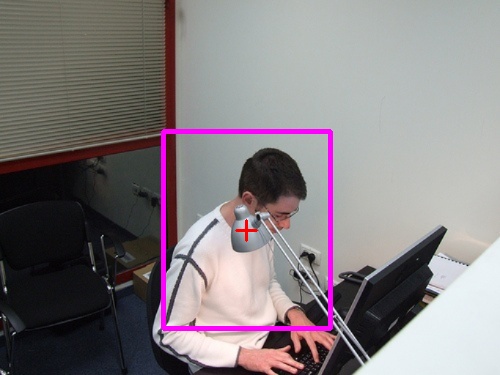}&
\includegraphics[height=\figh]{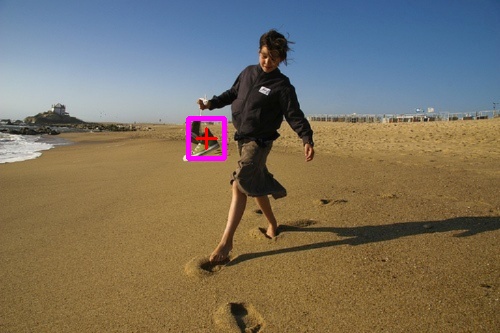}&
\includegraphics[height=\figh]{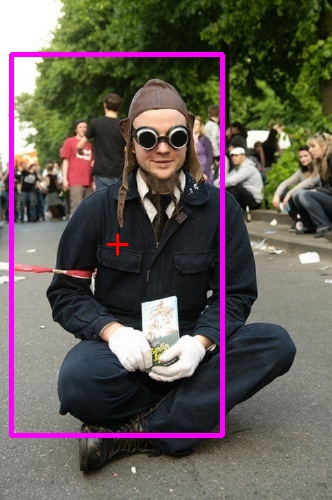}&
\includegraphics[height=\figh]{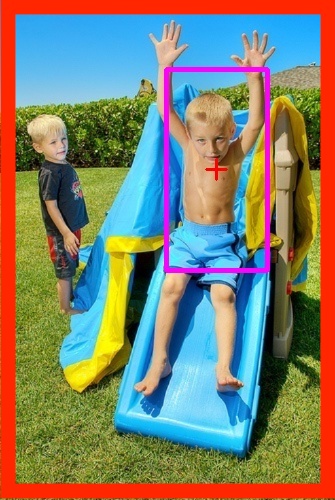}
\end{tabular}

\caption{ (Top) \oursspace heatmaps; (Bottom) IG \cite{gupta2020contrastive} predicted boxes; (Middle text) grounding queries. We show cases where \ours{} handles phrases which are less familiar or ambiguous to the detector. On the right, where the query is ambiguous, both methods failed.}
% \caption{ (Top) \oursspace heatmaps; (Bottom) IG \cite{gupta2020contrastive} predicted boxes; (Middle text) grounding queries. We show cases where \ours{} can help for objects that are less familiar or ambiguous to the detector. On the right, where the query is ambiguous, is a failure case for both methods.}
\figvspace
\label{fig:qualitative}
\end{center}
\end{figure*}

%% file: files/ablations.tex
\secvspace
\subsection{Ablations}\label{sec:ablations}
\secvspace
We used the Flickr30K \ourstask{} benchmark \cite{plummer2015flickr30k} with the `pointing game' accuracy measure \cite{zhang2016top} for analyzing the relative contribution and importance of the different components of our \ours{} approach. All the ablation studies were carried out on (the lighter) VGG backbone trained on MS-COCO, and using our complete \ours{} approach with all of its components except the ones being examined in each respective ablation sub-section below.

\secvspace
\subsubsection{Regularization losses $\mathcal{L}_{adv}$ and 
$\mathcal{L}_{neg}$}\label{sec:abl_reg}
\secvspace
\input{tables/regularization_ablation}

In Table \ref{tab:regularization_ablation} we evaluate the relative effect of our main regularization losses, namely: (i) the unconditioned adversarial loss $\mathcal{L}_{adv}$ responsible for suppressing the model parameters capitalizing on the artifacts of the synthetic blending we use for training our \oursspace models; and (ii) the negative text loss $\mathcal{L}_{neg}$ that drives to the empty heatmap output once the text is unrelated to the image. As we can see from the table, each of these regularization losses adds above $3.5\%$ to our \oursspace model performance affirming the benefits of their function. Moreover, when used jointly these two losses add more than $9\%$ to the overall accuracy.

\secvspace
\subsubsection{Image-to-text loss $\mathcal{L}_{i2t}$}\label{sec:abl_i2t_reg}
\secvspace
\input{tables/image2text_ablation}
Table \ref{tab:image2text} evaluates the benefit of the direct image-to-text matching loss $\mathcal{L}_{i2t}$ that is intended to improve the text and image features distributions alignment in order to facilitate better output of the conditioning module $\mathcal{C}$, as well as of an additional direct image-to-text attention output $\mathcal{H}_{i2t}$ resulting in the process of $\mathcal{L}_{i2t}$ computation. As the first row of Table \ref{tab:image2text} shows, training using only the $\mathcal{L}_{i2t}$ loss and using its corresponding output $\mathcal{H}_{i2t}$ (the only one available in this case) for grounding at test time is not sufficient for obtaining high performance. Significantly better accuracy (by almost $7\%$) is obtained via training using the $\mathcal{H}_{\ours}$ output and the \oursspace losses alone (without $\mathcal{L}_{i2t}$, second row). This indicates that the \oursspace losses contribute the most to the overall best result of $72.6\%$ attained when using all the losses and outputs jointly (third row). We believe that the reason for this might be that the \oursspace losses employ a (synthetic) structured training signal (learning to predict localized masked regions of the blended image conditioned on the text), while $\mathcal{L}_{i2t}$ loss capitalizes on unstructured (bag-of-words like) contrastive (in the batch) text to image matching.

\secvspace
\subsubsection{Blending alpha-map generation schemes}\label{sec:abl_alpha}
\secvspace
\input{tables/alpha_ablation_supp}
Table~\ref{tab:alpha_ablation_supp} evaluates some choices for the blending alpha-map ($\alpha$) generation scheme. Specifically, we examine the following alpha-map generators and their combinations (in portions of the batch): (i) the Perlin engine \cite{perlin1985image}; (ii) normalized pixel-wise combination of two random Gaussians: $\mathcal{G}\left[(x,y)|\mu_1,\sigma_1\right] / \sum_{j=1,2}\mathcal{G}\left[(x,y)|\mu_j,\sigma_j\right]$, with $\mu_j$, and $\sigma_j$ chosen at random and $\mathcal{G}$ being a Gaussian distribution; (iii) the Circle; and (iv) the Scale\&Shift. The Circle refers to a binary circular mask (with randomly generated center and radius), and the Scale\&Shift refers to a random scale and random relative shift blending of one of the images of the blended pair into the other image of the pair.
% Evaluation of more $\alpha$ generation schemes, including Gaussian mixtures and random shifting and scaling one image into another, is provided in supplementary. 
%
% As can be seen, the choice of the $\alpha$ generator has a non-negligible effect, varying the resulting model's accuracy in a range of about $4\%$. 
% We can also see that the best result naturally obtained when using a mix of different $\alpha$'s thus increasing their diversity.
We observed that mixing different alpha-map generation schemes in most cases leads to increased performance compared to each of the schemes alone, likely due to increased diversity of the mix. Mixing the Perlin and Gaussian schemes attains the best result.

\secvspace
\subsubsection{Language model ablation}\label{sec:abl_lang}
In Table \ref{tab:lang_model_supp} we evaluate the effect of the choice of the language model (ELMO \cite{peters2018deep} or BERT \cite{devlin2018bert}) used for the text embedding. 
% As expected, the more recent model (BERT) performs better in most cases, correspondingly to its original language tasks advantage. 
Notably, even with the ELMO text embedding our proposed \oursspace approach retains significant performance gains (between $1.1\%$ and $6.3\%$) above the results of \cite{akbari2019multi} for the corresponding VGG backbone (winning in all cases except when testing on ReferIt after training on VG).
\input{tables/elmo_ablation_supp}

\secvspace
\subsubsection{Attenuation for the text conditioning in $\mathcal{C}$}\label{sec:abl_cond}
\secvspace
\input{tables/condition_ablation}
In Table \ref{tab:cond_ablation} we evaluate several options for the type of the attenuation operation used in the conditioning module $\mathcal{C}$:\\ 1. \textit{Distance} attaining the best result stands for the attenuation described by eq. \eqref{eq:atten}. \\
2. Alternatively, we also test the \textit{projection} attenuation:
\begin{equation}
    \mathcal{A}(E^i,W^i) = \cos_{+}\left(E^i,W^i\right) \cdot E^i.
\end{equation}
3. The \textit{attention} attenuation via using a self-attention block accepting the concatenated $E^i$ and $W^i$ (replicated to each pixel of $E^i$) and outputting a tensor of the same size as $E^i$. \\
4+5. Two 'scalar' attenuations that return a single channel tensors outputs, \textit{dist2Atten}:
\begin{equation}
    \mathcal{A}(E^i,W^i) = \exp(- \cos\left(E^i,W^i\right))
\end{equation}
and \textit{cosine}:
\begin{equation}
    \mathcal{A}(E^i,W^i) = \cos_{+}\left(E^i,W^i\right)
\end{equation}

% \secvspace
% \subsubsection{Language model}\label{sec:abl_lang}
% \secvspace
% In Table \ref{tab:lang_model} we evaluate the effect of the choice of the language model for the text embedding. As expected, stronger models perform better, correspondingly to their language tasks performance. 
% Notably, even with the ELMO text embedding model our proposed \oursspace approach retains a significant $6.9\%$ performance gain over the result of \cite{akbari2019multi} using the corresponding VGG backbone and MS-COCO training.
% \input{tables/text_model_ablation}

%% file: tables/regularization_ablation.tex
\begin{table}
\begin{center}
\begin{tabular}{ c c|c }
%  \hline
  $\mathcal{L}_{neg}$ & $\mathcal{L}_{adv}$  & Pointing Accuracy\\ 
 \hline
 - & - & 63.22 \\
%  \hline
 \checkmark & - & 66.8\\
%  \hline
 - & \checkmark & 66.9\\
%  \hline
 \checkmark & \checkmark & \textbf{72.6} \\
%  \hline
\end{tabular}
\end{center}
\caption{The effect of the regularization losses $\mathcal{L}_{adv}$ and $\mathcal{L}_{neg}$. 
% The evaluation was done using the ``pointing game" accuracy on Flickr30K. All the the networks where trained with BERT+VGG models.
}
\figvspace
\label{tab:regularization_ablation}
\end{table}

%% file: tables/image2text_ablation.tex
\begin{table}
\begin{center}
\begin{tabular}{ c c | c c c  }
%  \hline
 \toprule
 \multicolumn{2}{c}{ Loss } & \multicolumn{3}{c}{Pointing Accuracy}\\
 $\mathcal{L}_{i2t}$ &  \oursspace losses & $\mathcal{H}_{i2t}$ & $\mathcal{H}_{\ours}$ & Mean\\
 \midrule
 \checkmark & - & 62.8 & - & - \\
 - & \checkmark & - & 69.5 & -  \\
 \checkmark & \checkmark & 68.2 & 67.1 & \textbf{72.6}\\
 \bottomrule
\end{tabular}
\end{center}
\caption{ Different combinations of the \ittlossfull{} loss, \ours{} losses, and heatmap generation schemes. Here '\ours{} losses' refers to our combination of $\mathcal{L}_{sep}$, $\mathcal{L}_{adv}$, and $\mathcal{L}_{neg}$.}
% The evaluation was done using the ``pointing game" accuracy on Flickr30K. All the the networks where trained with BERT+VGG models.
\figvspace
\label{tab:image2text}
\end{table}

% \begin{table}
% \begin{center}
% \begin{tabular}{ c c | c c c | c }
% %  \hline
%  \toprule
%  \multicolumn{2}{c}{ Loss } & \multicolumn{3}{c}{ Output Heatmap } & \multirow{2}{*}{Hit Accuracy}\\
%  $\mathcal{L}_{i2t}$ &  \oursspace losses & \mathcal{H}_{i2t} & \mathcal{H}_{\ours} & Mean & \\ 
%  \midrule
%  \checkmark & - & \checkmark & - & - & 62.8 \\
%  - & \checkmark & - & \checkmark & - & 69.5 \\
%  \checkmark & \checkmark & \checkmark & - & - & 68.2 \\
%  \checkmark & \checkmark & -& \checkmark & - & 67.1 \\
%  \checkmark & \checkmark & - & - & \checkmark &\textbf{72.6} \\
 
%  \bottomrule
% \end{tabular}
% \end{center}
% \caption{ Different combinations of the \ittlossfull{} loss and \gbs{} loss and heatmap generation. Here \oursspace losses refers to our combination of $\mathcal{L}_{\ours}$, $\mathcal{L}_{adv}$, and $\mathcal{L}_{neg}$.
% % The evaluation was done using the ``pointing game" accuracy on Flickr30K. All the the networks where trained with BERT+VGG models.
% }
% \label{tab:image2word}
% \end{table}

%% file: tables/alpha_ablation_supp.tex
\begin{table}[h]
\begin{center}
\begin{tabular}{ ccccc }
%  \hline
% \toprule
%  Perlin & Gaussian & Binary & Circle & Scale\&Shift & Pointing Accuracy\\ 
%  \midrule
%  100\% & 0\% & 0\% & 0\% & 0\% & 68.37 \\
% %  \hline
% %  \hline
%  50\% & 50\% & 0\% & 0\% & 0\% & \textbf{72.6} \\
%  50\% & 0\% & 50\% & 0\% & 0\% &  69.0\\
%  50\% & 0\%& 0\% & 50\% & 0\% &  66.9\\
%  50\% & 0\%& 0\% & 0\% & 50\% &  68.8\\
%  0\% & 100\% & 0\% & 0\% & 0\% & 70.7\\
%  0\% & 50\% & 50\% &0\% & 0\% &   67.2\\
% 0\%& 50\% & 0\% & 50\% & 0\% & 65.8\\
%  0\% & 50\% & 0\%& 0\% & 50\% & 68.0\\
%  0\% & 0\% & 100\% & 0\% & 0\% & 66.6\\
%  0\% & 0\% & 50\% & 50\% & 0\% & 68.9\\
%  0\% & 0\% & 50\% & 0\% & 50\% & 68.2\\
%  0\% & 0\% & 0\% & 100\% & 0\% & 65.79\\
%  0\% & 0\% & 0\% & 50\% & 50\% & 69.6\\
%  0\% & 0\% & 0\% & 0\% & 100\% & 66.0 \\
 
 \toprule
 Perlin & Gaussian & Circle & Scale\&Shift & Pointing Acc.\\ 
 \midrule
 100\% & 0\% & 0\% & 0\% & 68.37 \\
 50\% & 50\% & 0\% & 0\% & \textbf{72.6} \\
%  50\% & 0\% & 0\% & 0\% &  69.0\\
 50\% & 0\%& 50\% & 0\% &  66.9\\
 50\% & 0\%& 0\% & 50\% &  68.8\\
 0\% & 100\% & 0\% & 0\% & 70.7\\
%  0\% & 50\% & 0\% & 0\% &   67.2\\
0\%& 50\% & 50\% & 0\% & 65.8\\
 0\% & 50\% & 0\% & 50\% & 68.0\\
%  0\% & 0\% & 0\% & 0\% & 66.6\\
%  0\% & 0\% & 50\% & 0\% & 68.9\\
%  0\% & 0\% & 0\% & 50\% & 68.2\\
 0\% & 0\% & 100\% & 0\% & 65.8\\
 0\% & 0\% & 50\% & 50\% & 69.6\\
 0\% & 0\% & 0\% & 100\% & 66.0 \\
\bottomrule
\end{tabular}
\end{center}
\caption{Comparison of different variants of blending alpha-map generation including their mix (in $\%$ of the batch size). 
}
\figvspace
\label{tab:alpha_ablation_supp}
\end{table}

%% file: tables/elmo_ablation_supp.tex
\begin{table}[h]
\begin{center}
\begin{tabular}{ lcccc }
%  \cline{4-6}
\toprule
 \multirow{2}{*}{Language Model} & \multirow{2}{*}{Training} & \multicolumn{3}{c}{ Pointing Accuracy}\\
   &  & VG & Flickr30K & ReferIt\\ 
 \midrule% \hline
 ELMO & VG & \textbf{53.65} & 66.43 & 52.90 \\
 BERT & VG & 53.40 & \textbf{70.48} & \textbf{59.44}\\
 \midrule
 ELMO & MS-COCO & 49.03 & 67.9 & 49.37 \\
 BERT & MS-COCO & \textbf{52.00} & \textbf{72.60} & \textbf{56.10} \\
%  \hline
\end{tabular}
\end{center}
\caption{Comparison of ELMO vs. BERT text encoders in our \oursspace model using the VGG backbone.
}
\figvspace
\label{tab:lang_model_supp}
\end{table}

%% file: tables/condition_ablation.tex
\begin{table}
\begin{center}
\begin{tabular}{ c| c}
% \hline
 Condition Method & Pointing Accuracy\\
 \hline
 Distance & \textbf{72.60}\\
 Attention & 69.70  \\
 Projecton &  69.54\\
 Dist2Atten & 69.43 \\
 Cosine & 65.57 
%  \hline
\end{tabular}
\end{center}
\caption{Comparison of conditioning attenuation variants.}
\figvspace
\label{tab:cond_ablation}
\end{table}

% \begin{table*}[t!]
% \begin{center}
% \begin{tabular}{ c| c c c c c }
% % \hline
%  Condition Method & Distance & Attention & Projecton & Dist2Atten & Cosine\\ 
%  \hline
%  Pointing Accuracy & \textbf{72.60} & 69.70 & 69.54 & 69.43 & 65.57 \\
% %  \hline
% \end{tabular}
% \end{center}
% \caption{Comparison of different conditioning variants evaluated using the ``pointing game" accuracy on Flickr30K. All the the networks where trained with BERT+VGG models.}
% \label{tab:cond_ablation}
% \end{table*}

%% file: files/conclusions.tex
\section{Conclusion}\label{sec:conclusions}
We have proposed a compositional approach for training text grounding models with weak (text only) supervision and without reliance on pre-trained detectors. In our \ours{} approach, the model is trained to revert a random (synthetic) composition (blending) of images, using texts (describing the images) as guidance for associating each blended image pixel with the correct image. 
This enables the use of our model to perform phrase grounding at test time, by treating (real) query images as a composition of two regions, only one of which is associated to the query phrase. In addition to the \oursfull (\ours) idea, we propose a specific architecture and a set of important regularization losses enabling our \ours{} approach to achieve a new SotA in \ourstaskfull{} (\ourstask{}). 
% Specifically, we attain significant, up to $8.5\%$ improvements in grounding (pointing game) accuracy on the popular Flickr30K, VisualGenome, and ReferIt benchmarks. 
In addition, we show that our approach is complementary to the \textit{detector-based} WSG by demonstrating
significant
% over $7\%$ 
improvement of the detector-based WSG accuracy SotA on Flickr30K when using our \ours{} model in a naive combination with a detector-based approach. Finally, in a comprehensive ablation study, we carefully examine the relative importance and contribution of our approach's components and losses, clearly showing the contributions of the novel \ours{} ideas to its success.

Interesting future work directions, which are beyond the scope of this work, include: adversarial optimization for the blending-alpha via back-propagation; recurrent generation of the blending alpha by applying the trained model to non-blended batch images and conditioning on random parts of associated text; exploring vision transformer backbones; and applications to multi-modal grounding outside the text domain (e.g. grounding of sound in still images).  